\RequirePackage{fix-cm}
\documentclass[smallextended]{svjour3}
\smartqed
\usepackage{graphicx}
\usepackage{multirow}

\journalname{Data Mining and Knowledge Discovery}

\begin{document}

\title{Multi-task learning of daily work and study round-trips from survey data}

\author{Mehdi Katranji \and
        Sami Kraiem \and
        Laurent Moalic \and 
        Guilhem Sanmarty \and
        Alexandre Caminada \and
        Fouad Hadj Selem
}

\institute{
            M. Katranji \at
            VEDECOM, 77 rue des Chantiers\\
            78000, Versailles, France\\
            Tel: +331-85-360163\\
            \email{mehdi.katranji@vedecom.fr}
        \and
            L. Moalic \at
            University of Upper Alsace - LMIA EA 3993\\
            68100, Mulhouse, France
        \and 
            A. Caminada \at
            University of Nice Sophia Antipolis\\
            06000, Nice, France
        \and
            S. Kraiem, G. Sanmarty and F. Hadj Selem \at
            VEDECOM, 77 rue des Chantiers\\
            78000, Versailles, France
}

\date{Received: date / Accepted: date}

\maketitle

\begin{abstract}
In this study, we present a machine learning approach to infer the worker and student mobility flows on daily basis from static censuses. The rapid urbanization has made the estimation of the human mobility flows a critical task for transportation and urban planners. The primary objective of this paper is to complete individuals' census data with working and studying trips, allowing its merging with other mobility data to better estimate the complete origin-destination matrices. Worker and student mobility flows are among the most weekly regular displacements and consequently generate road congestion problems. Estimating their round-trips eases the decision-making processes for local authorities. Worker and student censuses often contain home location, work places and educational institutions. We thus propose a neural network model that learns the temporal distribution of displacements from other mobility sources and tries to predict them on new censuses data. The inclusion of multi-task learning in our neural network results in a significant error rate control in comparison to single task learning.
\keywords{Origin-destination matrix updating/estimating \and regular mobility pattern \and machine learning}
\end{abstract}

\section*{Author's contributions}

\paragraph{Mehdi Katranji:} coding, data analysis, drafting manuscript, experimentation, mathematical analyses, study design

\paragraph{Sami Kraiem:} data acquisition, manuscript correction, data analysis

\paragraph{Laurent Moalic:} manuscript correction

\paragraph{Guilhem Sanmarty:} data acquisition, data analysis, manuscript correction

\paragraph{Alexandre Caminada:} manuscript correction

\paragraph{Fouad Hadj-Selem:} supervising

\section{Introduction}
\label{section:introduction}
Population mobility dynamics has long been an important issue for local authorities in order to understand the usage of their road and rail networks or public transportation infrastructures. Over the years a lot of mobility data providers have been referenced, with their benefits and their drawbacks. Nowadays, there is an avalanche number of devices which are engaged in generating a plethora of volume of geolocated data. These devices include mobile phone, GPS transmitter/receiver, public transport smart cards or credit cards etc. Eventually, one can observe that several research work are devoted to the exploitation of those passive markers of mobility. However, none of these data sources are useful enough for exploitation and description of mobility flows in national proportion, with modal information, or travel purpose. Moreover, they lack trip-related and individuals' social attributes. This information is of great relevance to characterize the city structural dynamics, to estimate population needs in urban planning, transport demand and also for public health. For example, regarding age, keeping abreast with travel flux of younger and older population helps to optimize infrastructure such as location of schools, care facilities, etc.

Another aspect for which this information is relevant is the modelling of infectious diseases spreading \cite{lenormand2015influence}. To tackle that, Household Travel Surveys ($HTS$) are often used as control data when combining these different sources since they contain all the details on the demographic attributes, as well as on the transportation means, the travel purposes and the location of employment, housing and educational establishment. Furthermore, they offer a combination of highly detailed travel logs for carefully selected representative population samples. Nevertheless, they are expensive to administer and participate in. Thus, the time between each new survey is approximately 10 years in even the wealthiest cities \cite{cerema2012mobility}.

Fortunately, the datasets on the student and scholar mobility (MOBSCO, INSEE, 2014) and worker (MOBPRO, INSEE, 2014) \cite{insee2018website} are national censuses that aims to survey mobility over territories. The inherent features of such datasets make it among the most important mobility data source for mobility studies. In fact, those datasets are the widely available in terms of surveyed population, and it contains the modal information and socio-demographic attributes of the individuals. Unfortunately, they are no daily round-trip information contrarily to the $HTS$. In fact, no details on departure time nor arrival time are provided in this survey due to privacy concerns. Home and work/school related trips are generally done on very specific hours during the day and this specificity is an overwhelming issue for urban planning \cite{gonzalez2008understanding}. Indeed, these commuting patterns generate a lot of traffic and road congestion. Knowing the temporality of these regular trips is a key point to understand the human mobility and the decision-making process, which is quite useful for better management of cities.

The principal objective of this work is to estimate the temporal distribution of individuals described in the stationary census datasets by using a multi-task learning model \cite{zhou2012multi}. It allows to learn the temporal distribution of displacements from several old $HTS$, and tries to predict them on the MOBPRO and MOBSCO individuals. The model we propose has been validated in a real context.

\section{Related Work}
\label{section:related_work}
We noticed that the core topic of this study has been investigated from two dimensions while we examine the previous work. Hence, we shall elaborate both of them one by one in perspective of individuals' multi-task learning approach.

\subsection{Predictability of human mobility flows}
\label{subsection:predictability}

It is widely expected that strong regularities and similarities exist in human mobility patterns, thus we do believe that it is possible to learn such patterns from previous $HTS$ data and predict it using new census data. In fact, researchers have pointed out that individuals are predictable, unique and slow to explore new places. Precisely, it has been found in \cite{gonzalez2008understanding} that most individuals follow a simple and reproducible pattern regardless of time and distance. This causes a strong tendency to return to locations they visited before. Furthermore, most regular displacements are linked to home and work location \cite{arai2014understanding}. On the other hand, urban mobility is linked to socio-demographic characteristics and vary strongly with gender, age and occupation \cite{lenormand2015influence}. Besides, it also partially results from the interaction of socio-economic indicators similarly in all the countries \cite{louf2014scaling,louf2014congestion}. All these results suggest that it is scientifically possible to predict people's movement and that both individuals' attributes and socio-economic indicators of cities need to be used in the learning process. In \cite{arai2014understanding,arai2013estimation} the variables describing the demographic attributes of individuals are considered as the unknown target functions during the statistical learning to infer the human mobility patterns. We instead propose to use them as training data in order to estimate the temporal distribution of the displacements. We also propose not to limit ourselves to the attributes of the individuals but to use the socio-demographic statistics of the origin and destination cities as the points of interest description, unemployment rate or economic growth (see more examples later on). This is an important difference as it empowers the learning model to be generic outside the learning towns.

In \cite{katranji2016mobility} the authors used a machine learning technique to transfer the knowledge on the human mobility from one to another city using the same data as ours. Nevertheless, we can notice two main differences with our approach. First, the multi-task aspect in our model is more appropriate because the target variables are correlated. Thus, as we shall see later, performing multiple task simultaneously helps to capture intrinsic relatedness and so gets better results. Second, in this work we test a kind of deep architecture contrary to \cite{katranji2016mobility} in which only regular learning model are tested. As far as we know, this is the first work on such kind of data that employs a deep model. In fact, deep learning models are notoriously powerful on structured data (image, text, speech, …). Nevertheless, even if the hidden structure of urban mobility is not discovered yet, many interesting works have checked the scalability of the human mobility patterns across cities (see \cite{gonzalez2008understanding,arai2014understanding,lenormand2015influence,louf2014scaling,zheng2014urban}). 

In \cite{zhang2015comobile} the authors present an interesting approach to merge the Origin-Destination (O-D) matrices based on a multi-view learning framework called coMobile. They illustrate the improvement of the estimates due to the use of many sources together to overcome the problems and skews of uni-source models. Unfortunately, all these sources are temporally disaggregated and the suggested approaches do not allow to add census data. Hence, in this paper, we are motivated by the hope to enable including census data, that are aggregated, in order to be used in different fusion models. The multi-source models presented in \cite{zhang2015comobile} (see also \cite{zhang2014exploring} and examples in \cite{lenormand2014cross}) can also be enriched with census data once they are disaggregated as we propose in our method.

\subsection{Methodological aspects}
\label{subsection:methodological_aspects}

Deep learning has its roots in neural networks, which were originally inspired by the way neurons are connected in the brain \cite{hopfield1987neural}. Neural networks have been studied for over 60 years and a huge volume of literature exists on this topic. A neural network is typically used to learn a complex function between inputs (data) and outputs (response variables) in the absence of a model, so it can be thought of as a generalized regression framework. However, in the wake of its evolution, it is challenging to learn the weights of such a network followed by interpreting the hidden layers. So as learning problems became more complex, it was desirable to train networks with more hidden layers. Since their inception 30 years ago, deep architectures have proved to be adept at modelling multiple levels of abstraction \cite{hornik1991approximation}. However, they were notoriously difficult to train since their objective functions are non-convex and highly non-linear. Moreover, the level of non-linearity increases with the number of layers in the network raising difficulties in training large amount of data.

A major breakthrough was observed in 2006 when Hinton and Salakhutdinov \cite{hinton2006reducing} showed that a deep neural network can be trained effectively by first performing unsupervised pre-training one layer at a time, followed by supervised fine-tuning using a gradient-descent algorithm called back-propagation \cite{rumelhart1988learning}. The key point in this work was their clever way of initializing weights by training each layer one by one with unsupervised training. In this way, the obtained weights are much better than just giving them random values. This technique to separately train each layer is called a Restricted Boltzmann Machine (RBM). They applied their learning algorithm to dimensionality reduction of images and achieved substantially better results than conventional PCA-based methods. Nevertheless, so as of now, a lot of the top performing neural networks seem to be of a purely supervised nature. In fact, there have been various innovations in purely supervised training that have rendered unsupervised pre-training unnecessary.

One of the main innovations was moving away from sigmoidal (sigmoid, tanh) activation units, which can saturate and thus very little gradient gets propagated backwards, so learning is incredibly slow if not completely halted for all practical intents and purposes. The authors in \cite{glorot2011deep} used Rectified Linear Units (ReLUs) as activation functions instead of the traditional sigmoidal units. The ReLUs do not suffer from the vanishing gradient/saturating sigmoidal issues and can be used to train deeper nets. From that moment, many researchers noticed that using similar rectifying non-linearities seems to close much of the gap between purely supervised methods and unsupervised pre-trained methods.

Another innovation is that improved initializations have been figured out for deep networks. In fact, using the idea of standardizing variance across the layers of a network, the research community was successful in realizing and establishing de-facto rules of thumb in last decade. One of the first most popular work was \cite{glorot2010understanding} where they proposed a formula for estimating the standard deviation on the basis of the number of inputs and output channels of the layers under assumption of no non-linearity between layers. This formula is extended in \cite{he2015delving} to the ReLU non-linearity and showed its superior performance for ReLU based nets. From then, this Glorot initialization is frequently used and works well in many applications. So as of now, a lot of the top performing conv nets seem to be of a purely supervised nature.

One of the other big innovation has been the use of dropout training \cite{hinton2012improving}, which allows us to train deeper, bigger neural networks longer without overfitting as much. Dropout is a technique where randomly selected neurons are ignored during training. They are “dropped-out” randomly. This means that their contribution to the activation of downstream neurons is temporary removed on the forward pass and any weight updates are not applied to the neuron on the backward pass. The effect is that the network becomes less sensitive to the specific weights of neurons. This in turn results in a network that is capable of better generalization and is less likely to overfit the training data.

At this point, with all these discoveries since 2006, it had become clear that unsupervised pre-training is not essential to deep learning. It was helpful, without any doubt, but it was also shown that in some well-done cases, purely supervised training (with the correct starting weight scales and activation function) could outperform training that included the unsupervised step.

\section{Datasets}
\label{section:datasets}

The main goal of this work is to propose a model that transforms aggregated census data into disaggregated O-D trip matrices. This disaggregation is essential for the merging of such censuses with other types of mobility data \cite{lenormand2014cross}. In this work, three types of datasets have been used. The first two type are dedicated to the training and validating of the machine learning model while the third one is used for the application of the obtained model on the existing aggregated census data. 

\subsection{Household Travel Survey}

The first type of data we use include several conventional $HTS$. They are adjusted to the population of specific places and contains basic demographic attributes such as gender, age, job types, transportation means, origins, destinations, and purposes of displacements. Most of all, all these displacements are associated to a specific departure date and time. We propose to focus our application to home and scholar related displacements, thus, the surveyed trips from these $HTS$ are filtered to create a specific set of temporal mobility data. Our set of $HTS$ contains the results of the following official surveys: Arras (2014), Beziers (2014), Chalon-sur-Saone (2014), La-Roche-sur-Yon (2013), Longwy (2014), Lyon (2014), Montpellier (2014), Nantes (2015), Toulouse (2013) and EGT Parisian region (2010). Note that the total number of the used worker trips is around 35,000 and student trips around 22,000. Finally, as we use a machine learning model to infer the time distribution of worker and student related travels, we propose to split this global set of mobility data into two parts. A part is dedicated to the learning step of the model while the other part is used for the testing. We validate the results of our model on the testing part.

\subsection{Localized descriptive variables}

We use the second database during the training step and inferring step. It contains different descriptive variables of different cities in the studied areas. We propose to use several variables such as points of interest (POI), population size, the demography of the different socio-professional categories, the number of hospitals, malls, firms, etc. These data will serve as key variables for the learning process: we use them to transfer the "knowledge" of a source from a city to another. This is an open database published by the national institute of statistics and economic studies INSEE \cite{insee2018website} that collects, produces, and regularly disseminates information on the French economy and society. Similar open data are available in many other countries such as in USA via the United States Census Bureau or the Office of National Statistics in Great Britain.

\subsection{Census data}

The last datasets consist of the national mobility censuses called MOBPRO and MOBSCO. Both surveys are an open dataset published by the INSEE \cite{insee2018website}. They provide the residence, study or working location of almost 5 million students and 8 million workers (about one third of the population). Each individual is given a weight which corresponds to an adjustment factor. This surveys also contain the main socio-demographic characteristics. We propose to apply the validated results of our model to these existing data and to transform this aggregated censuses into O-D trip matrices.

\section{Methodology}
\label{section:methodology}

\begin{figure*}
\centering
\caption{Predictor and predicted variables use in model}
\label{figure:model}
\includegraphics[width=\textwidth]{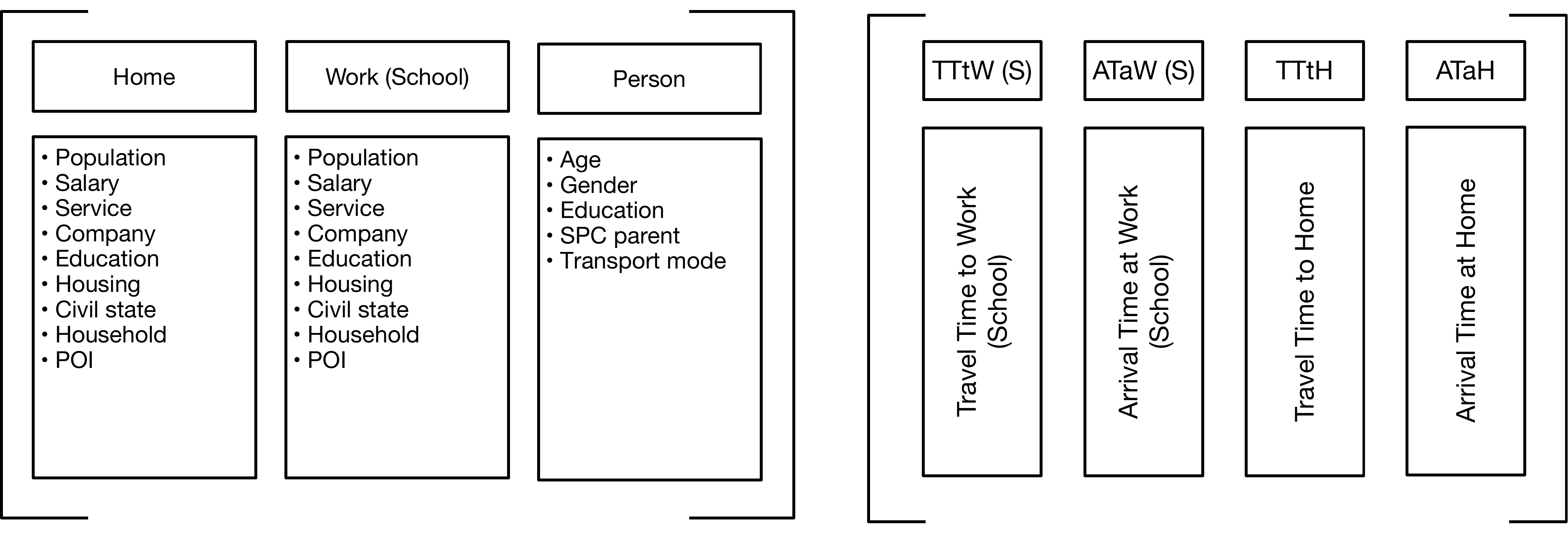}
\end{figure*}

\subsection{Data pre-processing}

First, let's precise the predictor and predicted variables for our problem before describing the pre-processing of the data. We select four target measures allowing to remodel the daily regular travel pattern namely: the Arrival Time at Work (ATaW), the Arrival Time at Home (ATaH), the Travel Time to Work (TTtW) and the Travel Time to Home (TTtH). Defined in this way, our O-D matrices  estimation problem is reduced to a classification of four related categorical targets:

\begin{enumerate}
 \item Arrival Time at Work (\textit{ATaW})
 \item Arrival Time at Home (\textit{ATaH})
 \item Travel Time to Work (\textit{TTtW})
 \item Travel Time to Home (\textit{TTtH})
\end{enumerate}

For convenience of our shrewd readers, we adopted the nomenclature such as "S" for School instead of "W". One can notice that the departure times to Work and to Home (respectively DTtW and DTtH) can easily be deduced from the previous information:

$$DTtW = ATaW - TTtW$$
$$DTtH = ATaH - TTtH$$

\begin{figure}
\caption{Sample of the four targets for workers (a) and students (b)}
\label{figure:sample_targets}
\begin{verbatim}

TTtW            ATaW            TTtH            ATaH       (a)
4:15 to 4:29    13:00 to 13:14  0:00 to 0:14    19:15 to 19:29
0:15 to 0:29    13:15 to 13:29  0:15 to 0:29    21:00 to 21:14  
5:15 to 5:29    13:15 to 13:29  0:15 to 0:29    15:15 to 15:29  
1:00 to 1:14    15:00 to 15:14  1:15 to 1:29    17:15 to 17:29  
0:00 to 0:14    6:00 to 6:14    0:00 to 0:14    18:15 to 18:29  

TTtS            ATaS            TTtH            ATaH       (b)
0:00 to 0:14    7:15 to 7:29    4:15 to 4:29    21:15 to 21:29
0:00 to 0:14    7:00 to 7:14    0:00 to 0:14    17:00 to 17:14
0:00 to 0:14    7:15 to 7:29    0:15 to 0:29    17:00 to 17:14
0:00 to 0:14    8:15 to 8:29    0:00 to 0:14    16:00 to 16:14
0:00 to 0:14    8:15 to 8:29    0:00 to 0:14    13:00 to 13:14
\end{verbatim}
\end{figure}

In particular, this estimate allows to generate the O-D matrices which also contain all the attributes inherited from the individuals' description of the census datasets. On the other hand, as illustrated in Fig.~\ref{figure:model}, the predictor variables are structured into three categories and theoretical travels times:

\begin{itemize}
    \item Urban context data of the home location town: it contains many statistics about points of interest and land use as service, company, housing, shopping, etc. These statistics are closely linked to mobility flows \cite{zheng2014urban}. In addition, it has been identified a strong correlation between this data and the traffic patterns including travel time  as well \cite{zheng2014urban}. Including these POI like data enables our model to be applicable outside the environment in which it has been optimized.

    \item Urban context data of the work or school location town (similar to the previous one).

    \item Theoretical travels times: theoretical estimation of travel duration between the two cities (couple origin-destination) using public transportation or personal vehicle. Obviously, they are useful predictor. Moreover, they have been collected using online map services (\cite{navitia2018website,graphhopper2018website}).

    \item Person attributes: it contains all the variables that may be useful for predicting travel habit such as gender, age, job types and transportation means. In fact, the HTS and census data have different dimensions, but with enough common variables to allow estimating the unique multivariate probability distribution that models the human mobility patterns. It contains all the variables that may be useful for predicting travel habits. These variables include gender, age, job types, salaries, grade levels, family status... Thus, these common variables are extracted from both sources to form this group of features.  
\end{itemize}

The selected displacements from the $HTS$ surveys are then grouped in a single matrix and the four corresponding targets (arrival times and travel durations) are also aligned in a second matrix. Finally, the learning matrix is completed by the other predictor variables using the origin and destination towns has shown in Fig.~\ref{figure:model}.

\subsection{Design of the learning model}

The target variables are real numbers and could be estimated using regression models. Nevertheless, it is not evident to estimate the arrival time accurately. Moreover, we perform a lot of tests based on regression techniques with popular models such as $SVM$, Elastic-Net \cite{pedregosa2011scikit} and the scores were always low. Hence, we proceed by classification. In fact, we split the 24 hours of the day into 96 time slots. Thus, we have transformed the four target values into categorical variables. Thereby, we seek the classification of different time values in these specific intervals. The time slots have been fixed manually (for instance, 08:00 to 8:15, 8:15 to 8:30, etc.). 

Defined in this way, our disaggregation problem is reduced to a classification of four categorical  targets as shown in Fig.~\ref{figure:sample_targets}. An important problem in machine learning is how to generalize between multiple related tasks. This problem has been called "multi-task learning", "learning to learn", or in some cases "predicting multivariate responses". Multi-task learning is an approach to learn a problem together with other related problems at the same time, with the aim of mutual benefit. The simplest approach can be a neural network or its advanced version known as deep learning since it is based on a set of hidden units that are shared among multiple tasks. Thus, we propose to tackle our problem using a multi-task neural network model. Recently, such deep learning based methods have made great progress in various fields and various innovations have been discovered as we have pointed out in \ref{subsection:methodological_aspects} for the sake of completeness and in order to justify our choices.

\subsection{Classification algorithm and hyperparameters selection}

\begin{figure}
\centering
\caption{Methodology overview}
\label{figure:methodology_nn}
\includegraphics[width=3.4in]{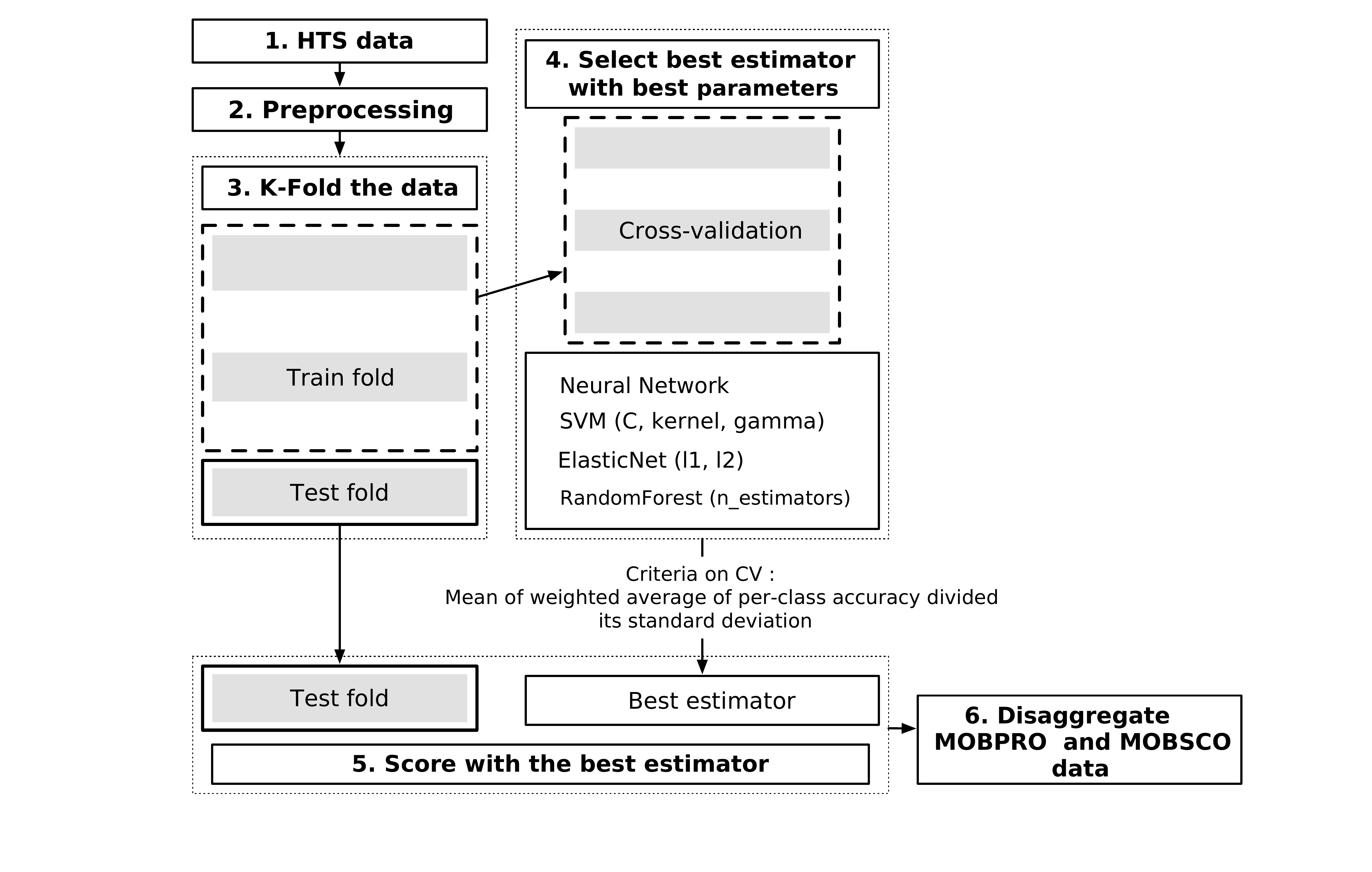}
\end{figure}

Following the previous section \ref{subsection:methodological_aspects}, we define our multi-task classification model by combining some popular neural networks styles and algorithms:

\begin{itemize}
    \item Glorot initialization \cite{he2015delving}.

    \item Dropout regularization \cite{hinton2012improving} with 0,25 or 50 percent rate : this hyper-parameters will be tuned using a nested cross validation loops as explained later on.

    \item Rectified Linear Unit activation function (ReLU) \cite{glorot2011deep}.

    \item Adam minimization algorithm \cite{kingma2014adam}.
\end{itemize}

All the experiments are run in a 10-fold cross-validation (CV) fashion, using the same folds for each approach, as shown in Fig.~\ref{figure:methodology_nn}. For each fold, the entire training process (hyper-parameters selection and fine tuning the neural network using Adam minimization algorithm) is repeated using only the training data in that fold before predictions are inferred on the test set of it, to ensure a fair evaluation. For all hyper-parameters selection, we tune the parameters in a nested CV fashion. In each training-testing experiment of the 10-fold CV, we have nine folds for training and one fold for testing. On the nine folds of training data, we carry out a nine-fold CV (eight folds for training and one fold for tuning) to select the best parameters.

Once the "best" subset has been identified using the inner cross-validation (using the best average of validation scores), the model is rebuilt from the full training set (the particular training set of the outer cross-validation) to get test scores on unseen data (which is used later for the selection and comparison tasks). Moreover looking for relevant and easily interpretable information, we use the scikit-learn weighted F1 score \cite{pedregosa2011scikit} for multi-class case to measure the classification performance for both inner and outer CV loops. This consists of computing the F1 scores for each label and then their average is weighted by support (which is the number of true instances for each label). Beside, since the 10 CV folds share 90\% of their training samples, no unbiased significance measure can be directly obtained from the obtained scores, Thus, we used permutation testing to asses empirical $p$-values. The null hypothesis was simulated by 1,000 random permutations of the four target variables, within each permuted sample we executed the whole 10 CV round. Then the statistics on the true data were compared with the ones obtained on the permuted ones. On the other hand, the number of epochs is selected, at each run, using a validation-based early stopping method \cite{prechelt1998early}. This allows us to avoid overfitting by stopping the training of the model as soon as the loss of its corresponding validation set stops decreasing. 

\begin{figure}
\centering
\caption{Uniform and pyramidal neural network structure}
\label{figure:uniform_vs_pyramidal}
\includegraphics[width=3.4in]{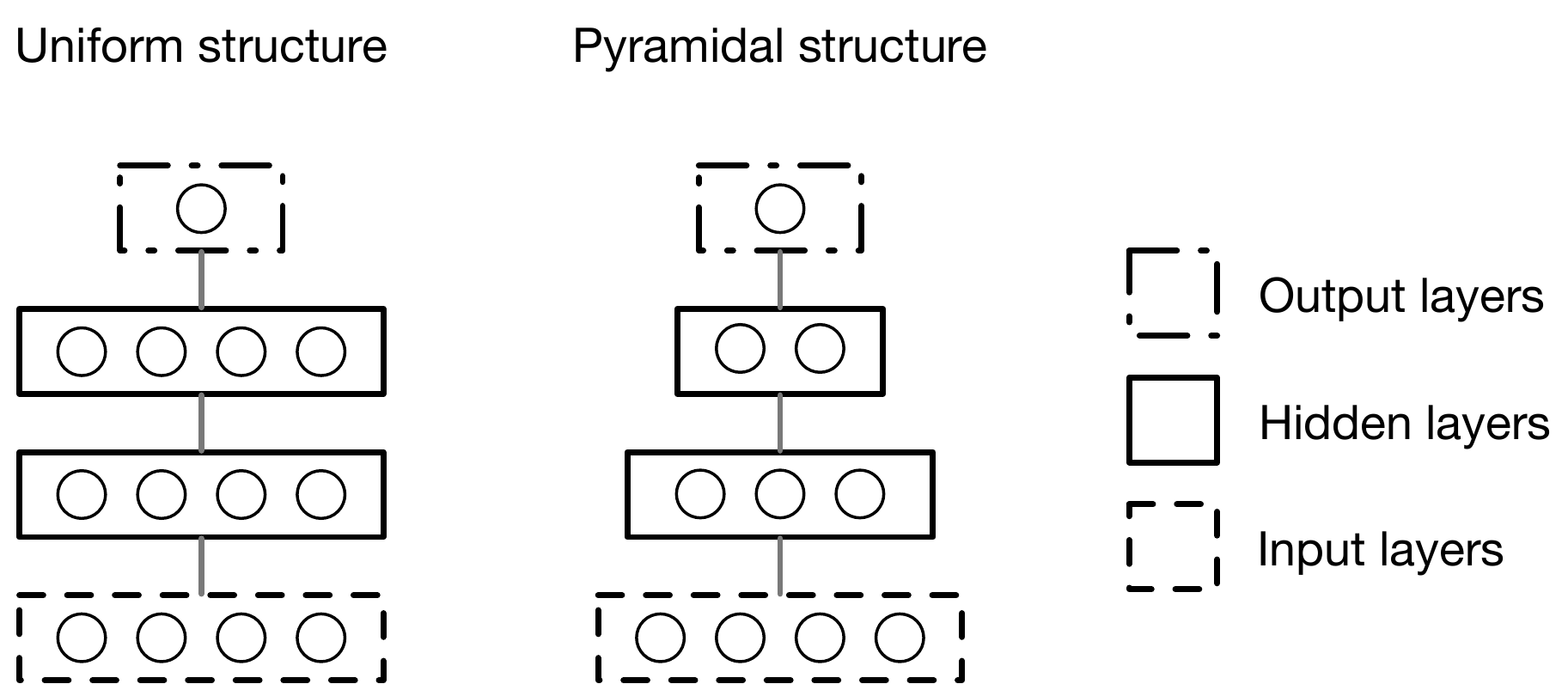}
\end{figure}

Now we address the issue of the network architecture selection. The main two points are the choice of the "better" number of hidden layers and of the hidden units in each layer. For this, we use the nested 10-fold cross-validation as described above for this model selection task. In fact, we vary the number of hidden layers from two to 10 and the network structure following two rules of thumb \cite{ding2008neural}. In that way, we tested the uniform structure with a fixed number of nodes (two-thirds of the sum of the input and output numbers) and the pyramidal structure in which the number of nodes is declining from the input layer to the output layer with fixed rate (thus the number of units in depends on the number of layer, Fig.~\ref{figure:uniform_vs_pyramidal}).

\begin{figure}
\centering
\caption{ATaW/ATaH mean classification score of multi-task NN (pyramidal structure) with different numbers of hidden layers}
\label{figure:layer}
\includegraphics[width=3.8in]{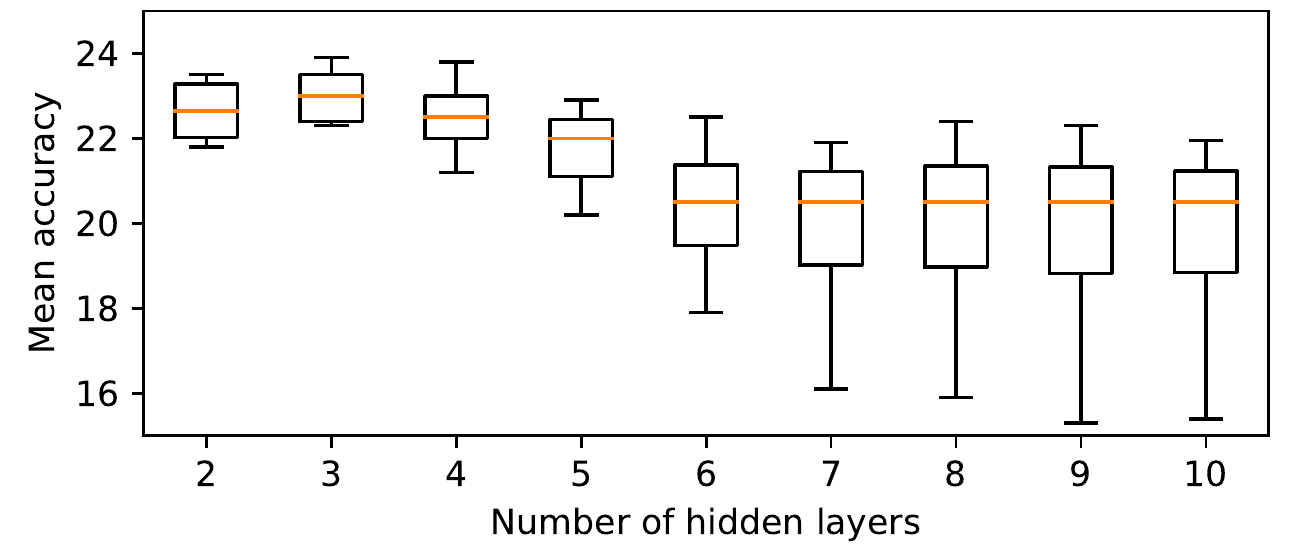}
\end{figure}

In this way we get 18 different neural network structures. For each model, we only tune one hyper-parameter which is dropout training rate \cite{hinton2012improving} across three values (0, 0.25 and 0.5) using the nested cross validation method as describes above. The comparison is made based on the average of the obtained 10 test scores for the four targets (the average of 40 values). An example of this computation is shown in Fig.~\ref{figure:layer} for the 10 pyramidal structure models which seems to perform better in terms of goodness and stability of the obtained score. It can be pointed out in Fig.~\ref{figure:layer} that using only three hidden layers performs higher score with low variance. Finally, we have concluded  that the winning model is a variant of a Multi-Layer Perceptron (MLP) with pyramidal architecture and three hidden layers.

\subsection{Single-task versus multi-task learning}

\begin{figure}
\centering
\caption{Test scores of the 4 target values using multi-task (blue) vs a single-task (green) worker models}
\label{figure:multitask}
\includegraphics[width=3in]{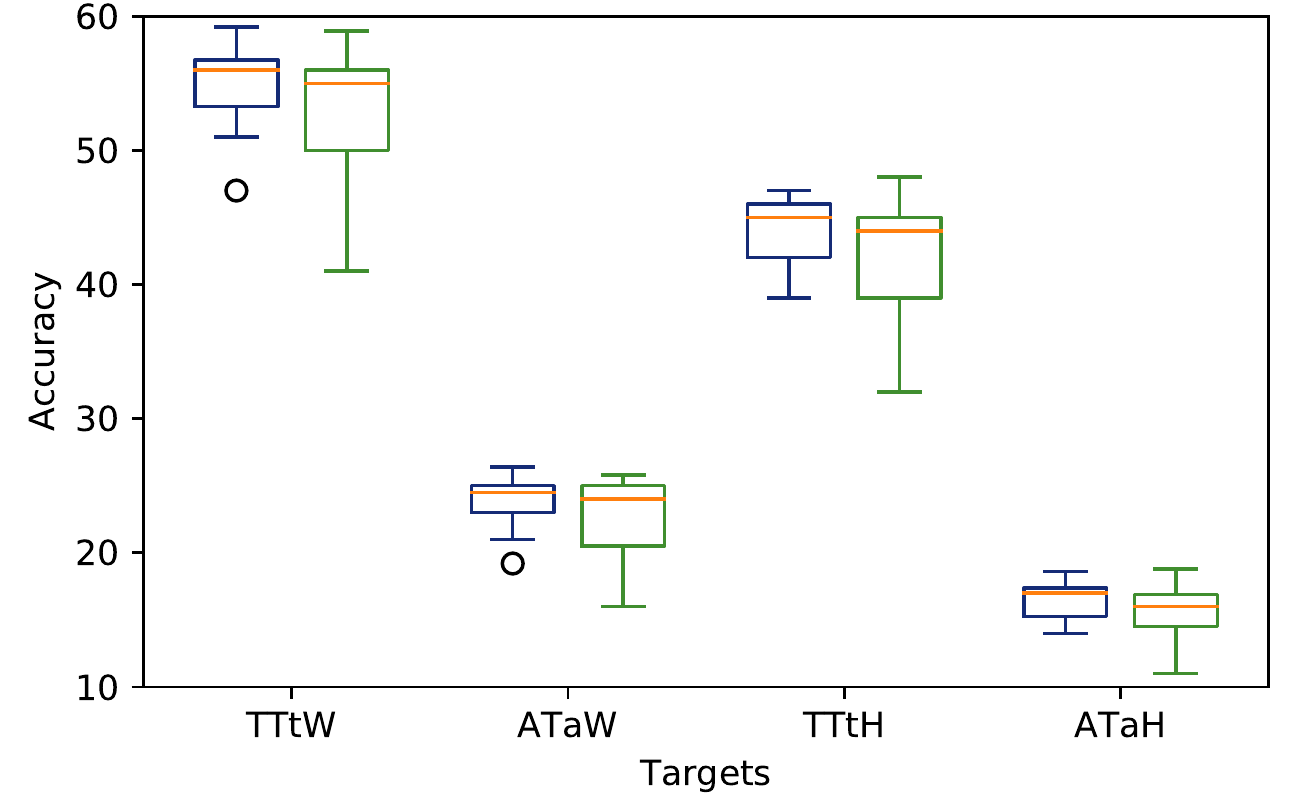}
\end{figure}

The purpose of this paragraph is to compare single and multi-task learning specificity to determine whether the joint interference of the fours targets values is beneficial for predicting each single variable. To this end, we have used the same 18 neural network models but in a single tasks fashion (each target is inferred separately). The resulting models are then compared the 18 multi-task ones using the same 10 cross-validation techniques based on the test scores. We notice that this comparison is not limited to the winning model with pyramidal architecture and two hidden layers. This allows to examine the impact of the multi-task approach in an unbiased way. It can be pointed out in Fig.~\ref{figure:multitask} that the multi-task models outperform the single-task ones for all variables. It suggests that the joint interference is beneficial to our prediction problem not only in terms of correct classification score but also in stability. This reflects the strong correlation between the four targets that may be better considered using shared hidden units among tasks. Moreover, the single task models may lead to very low scores for some unsuitable versions contrary to the multi-task case in with the obtained scores are generally stable.

\section{Results}
\label{section:results}

\subsection{Main observations}

In Table~\ref{table:accuracy:workers} we show a summary of our prediction results for used models. The first and second rows show the validation and test scores for each target. We evaluate average scores from selected hyper-parameters with the best validation F1 score over all runs and evaluated it on the test set. In the third row we show the average absolute error measured in minutes for each target. That is, when an individual is misclassified, how far is (in average) the results provided by the underlying model, from the correct value. It can be pointed out that:

\begin{itemize}
     \item Neural-network based model improves the test scores for all targets by 5, 5, 8 and 0 points on average compared to the purely linear model $SVM$. Moreover, it always reduces the difference between test and validation scores by 6, 12, 9 and 10 points). Finally, it reduces the average error value by 1.7, 5.9, 3.5 and 6.7 minutes. We conclude that, multi-task neural-network allows us a higher prediction score with smaller error and better generalization performance. In fact, a large gap between validation and test scores denote over-fitting that prevents good generalization. 
     
    \item $ReLU$ + Glorot outperforms $SVM$ by improving the test scores, reducing the difference between test and validation scores and reducing the average error value for the four targets.
    
    \item These are not very high scores but still statistically significant due to the high number of classes.
    
    \item The validation scores for the second and fourth targets are low. We have checked that when increasing the number of epochs, we can increase the training score till a very high scores (about 95\%). Nevertheless, the validation scores start decreasing at the obtained low values. This suggests that we need to increase the complexity of our model by adding some hidden layers (or units in some layers). This is marked for future work.
    
    \item  The error values by the $ReLU$ + Glorot based model suggest that when individuals are misclassified, the estimated values are never significantly wrong (34.4 minutes on average). It signifies that the aggregated O-D matrices provided by the proposed model are close to the reality.
\end{itemize}

\begin{table}[t]
\caption{Performance scores for home-work related displacements. Significance notations: ***:~$p\leq10^{-3}$, **:~$p\leq10^{-2}$, *:~$p\leq0.05$. Empirical $p$-values were assessed by $1\,000$ executions of the 5CV round on randomly permuted data.}
\label{table:accuracy:workers}
\begin{center}
    \begin{tabular}{|c|c|c|c|c|}
    \multicolumn{5}{c}{\textbf{ReLU + Glorot}} \\\cline{1-5}
    Target                      & TTtW              & ATaW              & TTtH              & ATaH\\
    \hline
    Validation (in \%)          & $65$              & $38$              & $56$              & $26$\\
    \hline
    Test (in \%)                & $59\pm2.1^{***}$  & $26\pm2.0^{***}$  & $47\pm2.3^{***}$  & $16\pm1.9^{**}$\\
    \hline
    Error (in minutes)          & $13.5 $           & $42.5$            & $26.2$            & $55.4$\\
    \hline
    \multicolumn{5}{l}{\textbf{ }}\\
    \multicolumn{5}{c}{\textbf{Single-task SVM}} \\\cline{1-5}
    Target                      & TTtW              & ATaW              & TTtH              & ATaH\\
    \hline
    Validation (in \%)          & $68$              & $32$              & $52$              & $22$\\
    \hline
    Test (in \%)                & $53\pm1.9^{***}$  & $21\pm1.3^{**}$   & $39\pm1.8^{***}$  & $16\pm2.0^{**}$\\
    \hline
    Error (in minutes)          & $15.2 $           & $48.4$            & $29.7$            & $62.1$\\
    \hline
    \end{tabular}
\end{center}
\end{table}

In Table~\ref{table:accuracy:students} we show a summary of our prediction results using the weighted F1 score. The first column shows that the TTtS and ATaS target values are correctly determined in nearly 53\% contrary to TTtH and ATaH values for which the computed score is about 38\%. These are not very high scores but still statistically significant due to the high number of classes (see also the simulated p-values). We also notice  that gap between validation and test scores is acceptable (less than 6 $\%$ on average) which suggests a good generalization behavior. On the other hand, these results show only a partial view of how the model works. Indeed, it seems important to analyze what is happening when individuals are misfiled. The Table~\ref{table:accuracy:students} provides such an analysis. It shows the average absolute error measured in minutes for each target. As we can see, the durations are not significantly wrong, only few minutes in the first three cases. It means that even if an individual is misfiled, in average the results provided by the proposed model are close to the reality in most cases. For the arrival times at home, the average error is about 44 minutes. In fact, it seems that the ATaH is harder to predict because the higher the student grade are, the more scattered exit of school are.

\begin{table}[t]
\caption{Performance scores for home-study related displacements. Significance notations: ***:~$p\leq10^{-3}$, **:~$p\leq10^{-2}$, *:~$p\leq0.05$. Empirical $p$-values were assessed by $1\,000$ executions of the 5CV round on randomly permuted data.}
\label{table:accuracy:students}
\begin{center}
    \begin{tabular}{|c|c|c|c|c|}
    \multicolumn{5}{c}{\textbf{ReLU + Glorot}} \\\cline{1-5}
    Target                      & TTtS              & ATaS              & TTtH              & ATaH\\
    \hline
    Validation (in \%)          & $60$              & $60$              & $45$              & $42$\\
    \hline
    Test (in \%)                & $55\pm1.6^{***}$  & $52\pm1.0^{***}$  & $41\pm1.4^{***}$  & $35\pm1.8^{***}$\\
    \hline
    Error (in minutes)          & $8.2$             & $16.2$            & $14.3$            & $44.2$\\
    \hline
    \multicolumn{5}{l}{\textbf{ }}\\
    \multicolumn{5}{c}{\textbf{Single-task SVM}} \\\cline{1-5}
    Target                      & TTtS              & ATaS              & TTtH              & ATaH\\
    \hline
    Validation (in \%)          & $68$              & $63$              & $53$              & $44$\\
    \hline
    Test (in \%)                & $56\pm2.9^{***}$  & $51\pm2.3^{**}$   & $39\pm3.8^{**}$   & $31\pm2.1^{**}$\\
    \hline
    Error (in minutes)          & $8.3$             & $24.4$            & $23.1$            & $53.8$\\
    \hline
    \end{tabular}
\end{center}
\end{table}

\subsection{Portability to other places}
\label{subsection:portability}

A model, for being relevant, has to be replicable. This is what we propose to study in this paragraph. Up to now, the training of the proposed learning model was done on a subset of each $HTS$, and the validation was done on the not used part of these surveys. Now, we propose to do the training on all the surveys except one, and to apply the resulting model on the last and not used survey. For this end, we compute in each case (survey) the variation of the obtained classification score with respect to the reference scores in (Table~\ref{table:accuracy:workers},~\ref{table:accuracy:students}).

Table \ref{table:portability:workers} shows these results for each studied city, applied on the arrival time at work (ATaW) and arrival time at home (ATaH) variables. We have examined that the two other targets behave similarly. The first important thing we can notice is that removing the studied city from the training and validation sets has no significant effect (around only -2\%, -6\% in the worst case). This is a healthy sign about the application of our model in a real context in a given city particularly when no old $HTS$ data is available. The second observation is that removing Paris of the training step, which represents 31 $\%$ of the interviewees, leads to the worst case (-6$\%$ and -3$\%$). This shows the importance of the number of interviewees. This should also be the case for Toulouse and Montpelier. However, we notice that removing the Roche-sur-Yon survey (only 2.9$\%$ of the total number of trips) also leads to worse result (loss of 5$\%$ and 4 $\%$). We think that this is linked to its small size and the particular transport network (there are no trains, trams and metros).

\begin{table}[h]
\caption{Worker model portability scores, excluding one HTS survey from the learning process and test on it (HTS\_excluded\_score - ref\_score)}
\label{table:portability:workers}
\begin{center}
    \begin{tabular}{|c|c|c|c|c|c|}
    \hline
    \textbf{Exclude one (\%)} & \textbf{ATaW} & \textbf{ATaH}              & \textbf{Exclude one (\%)}      & \textbf{ATaW}  & \textbf{ATaH}\\
    \hline
    Arras (2.2)         & $+00$ & $-02$             & Lyon (14.8)               & $+02$ & $-04$\\
    \hline
    Beziers  (2.8)      & $-02$ & $-02$             & Montpellier (9.4)         & $-04$ & $-01$\\
    \hline
    Chalon (2.2)        & $+00$ & $-01$             & Nantes (17)               & $-01$ & $-01$\\
    \hline
    Roche (2.9)         & $-05$ & $-04$             & Toulouse  (11)            & $-04$ & $-01$\\
    \hline
    Longwy (1.8)        & $-02$ & $-02$             & Valence (4.8)             & $+03$ & $+00$\\
    \hline
    Paris (31)          & $-06$ & $-03$             & \textbf{Reference (100)}  & $26$  & $16$ \\
    \hline
    \end{tabular}
\end{center}
\end{table}

Table~\ref{table:portability:students} shows these results for our student model, applied on the ATaH and ATaS targets. We have checked that the two other targets behave similarly. The first important thing we can notice is that removing the studied city from the training set has positive effect in 4 cases (Chalon-sur-Saone, Longwy, Lyon, Montpellier) and no significant effect in the other two cases. This clearly proves that the model can be relevantly applied on another city. Moreover, the improvement of the classification score is linked to the homogeneity of the validation data (the validation data contains only the not used survey). This is a healthy sign about the application of our model in a real context in a given (one) city. The second observation is that removing Ile-de-France of the training, which still represents 31\% of the interviewees, also leads to the worst case like our worker model. That shows the importance of the number of interviewees in the quality of the results. Finally, we notice that removing the Arras survey also leads to worse result (loss of 3\%). We think that this is linked its small size and particular transport network (does not contain any transit service). In this computation, the training data only contains big cities.

\begin{table}[h]
\caption{Student model portability scores, excluding one HTS survey from the learning process and test on it (HTS\_excluded\_score - ref\_score)}
\label{table:portability:students}
\begin{center}
    \begin{tabular}{|c|c|c|c|c|c|}
    \hline
    \textbf{Exclude one (\%)} & \textbf{ATaS} & \textbf{ATaH}               & \textbf{Exclude one (\%)}      & \textbf{ATaS}  & \textbf{ATaH}\\
    \hline
    Arras (2.2)         & $-02$ & $-03$             & Lyon (14.8)               & $+01$ & $+06$\\
    \hline
    Beziers  (2.8)      & $-02$ & $-02$             & Montpellier (9.4)         & $+00$ & $+01$\\
    \hline
    Chalon (2.2)        & $+02$ & $+03$             & Nantes (17)               & $-02$ & $-01$\\
    \hline
    Roche (2.9)         & $-03$ & $+01$             & Toulouse  (11)            & $-01$ & $+02$\\
    \hline
    Longwy (1.8)        & $+04$ & $+04$             & Valence (4.8)             & $-01$ & $-01$\\
    \hline
    Paris (31)          & $-09$ & $-05$             & \textbf{Reference (100)}  & $52$  & $35$ \\
    \hline
    \end{tabular}
\end{center}
\end{table}

\subsection{Inferring Origin-Destination matrix}

Now, we describe the disaggregation of the MOBPRO and MOBSCO datasets. The main step consists of predicting the temporality of the displacements of each individual in the census data using the best selected model in the training step. We also keep the target $HTS$ area out of the training step, like in \ref{subsection:portability}. Next we build a histogram for each O-D pair and use a kernel density estimation approach to approximate the probability density function in each case. It is worth noticing that the mean error is very useful to the bandwidth parameter selection. It results in a global estimation of the scholar displacement between each two towns with a high precision level.

\begin{figure}
\centering
\caption{Comparison of student and worker mobility flows from HTS real data and inferred disaggregated\-census by our model}
\label{figure:comparison_hts}
\includegraphics[width=0.49\textwidth]{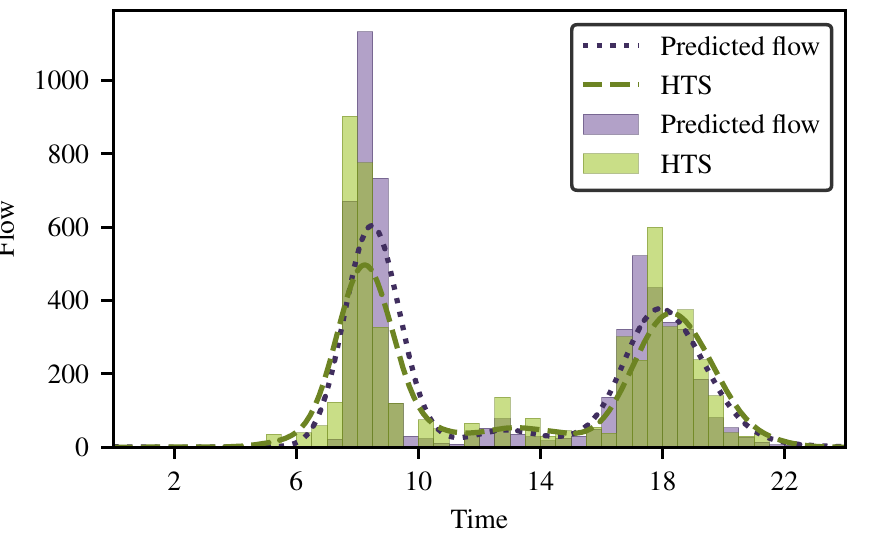}
\includegraphics[width=0.49\textwidth]{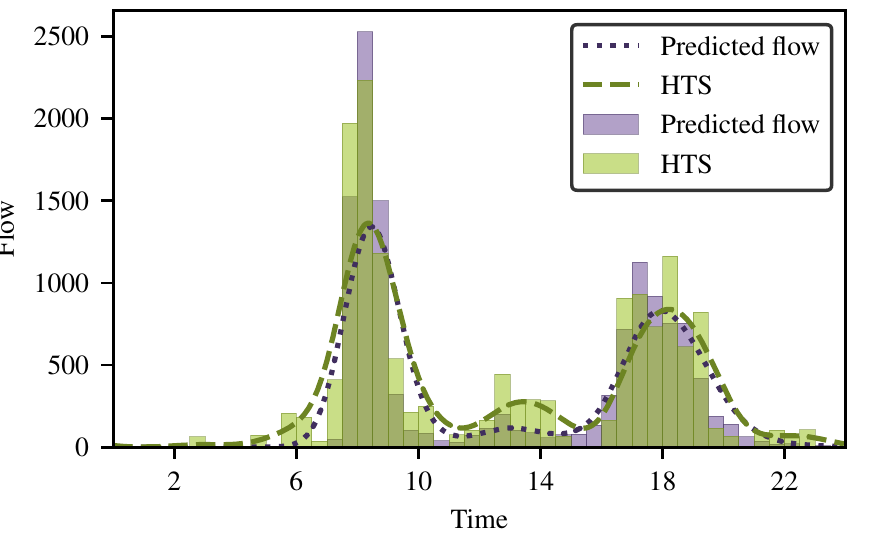}
\includegraphics[width=0.49\textwidth]{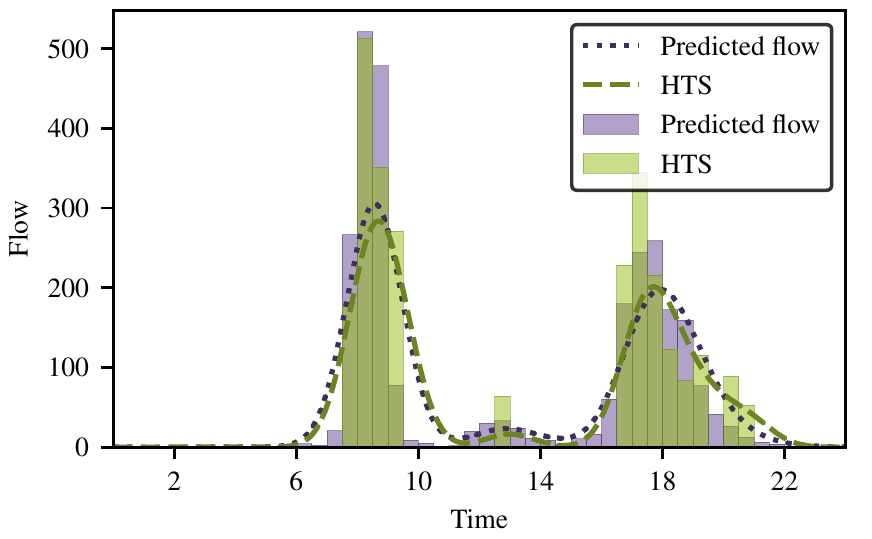}
\includegraphics[width=0.49\textwidth]{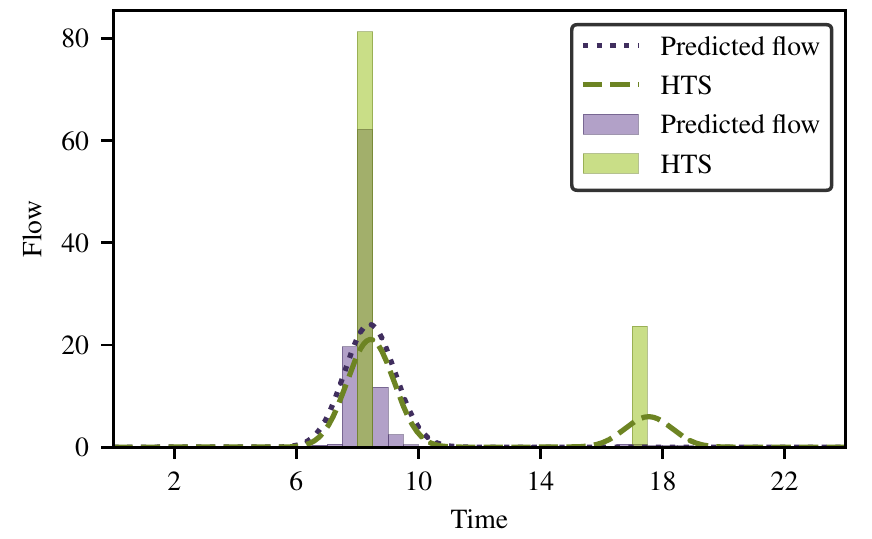}
\includegraphics[width=0.49\textwidth]{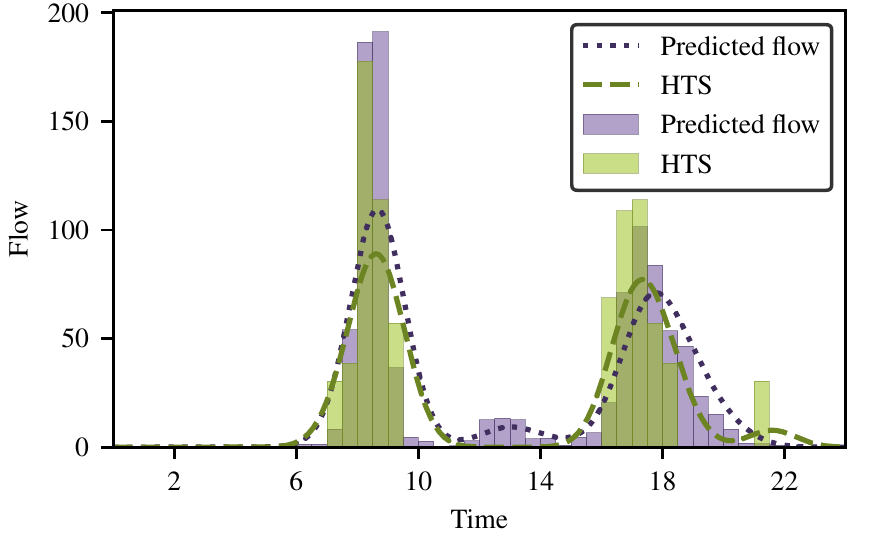}
\includegraphics[width=0.49\textwidth]{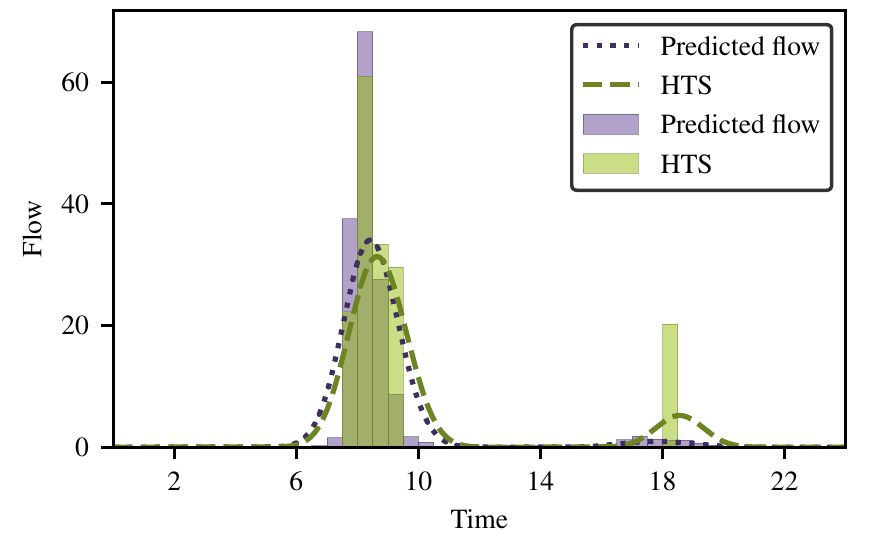}
\includegraphics[width=0.49\textwidth]{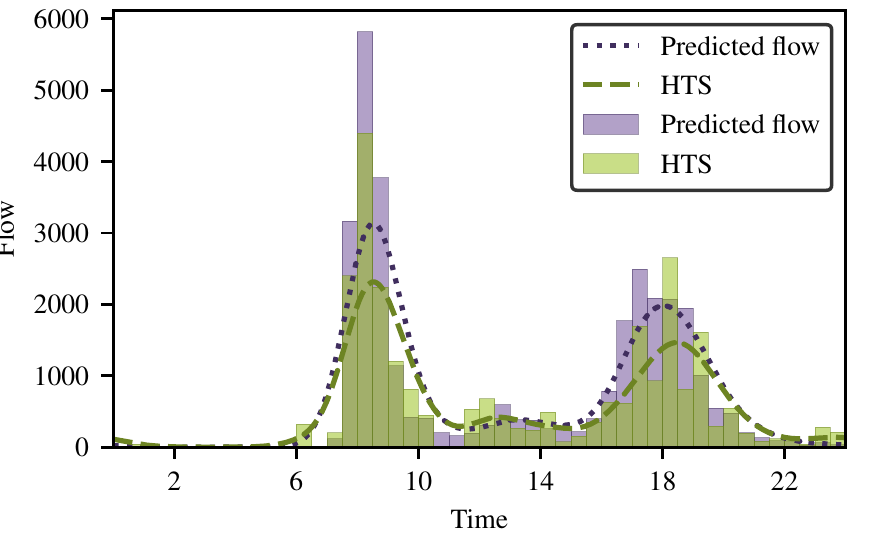}
\includegraphics[width=0.49\textwidth]{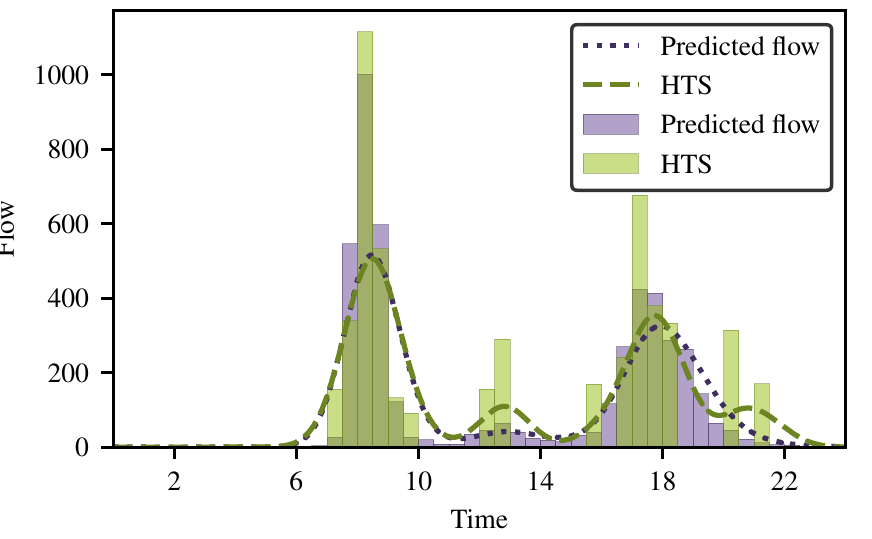}
\includegraphics[width=0.49\textwidth]{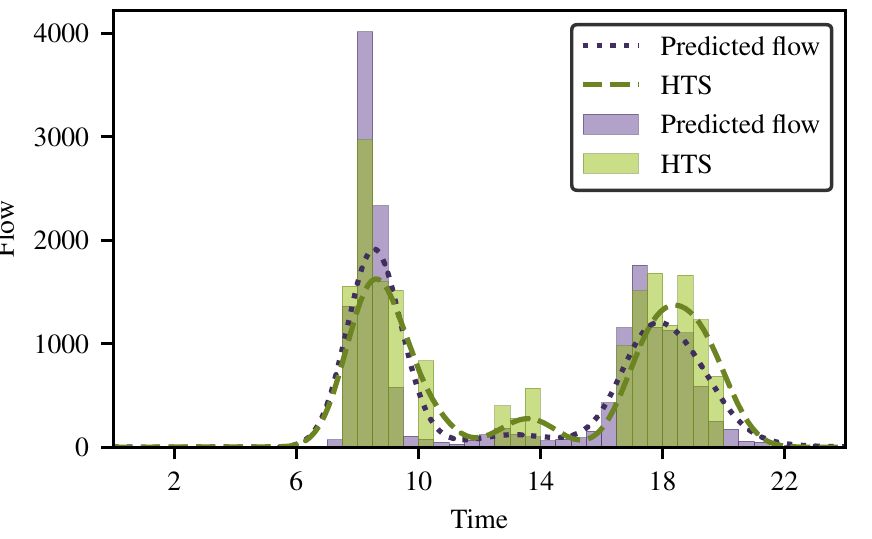}
\includegraphics[width=0.49\textwidth]{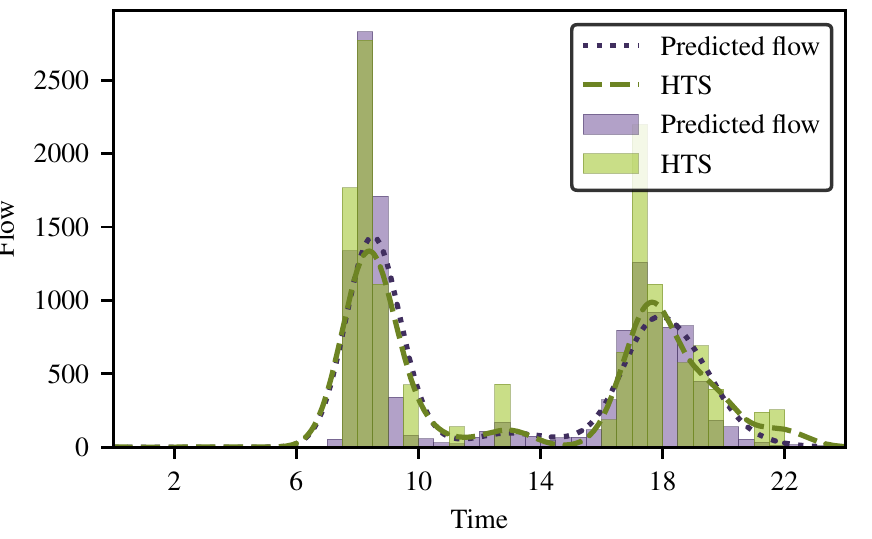}
\end{figure}

An example of such calculation is illustrated in Fig.~\ref{figure:comparison_hts}. It displays the estimated histogram by our approach together with a close-to-reality O-D matrix obtained from the our real HTS surveys. This plot shows a good agreement between the O-D matrices simulated using real HTS dataset and our census disaggregated data. Furthermore, we notice that the work and study round-trip flows are strongly linked to classical commuting periods (08:00 to 10:00 and 16:00 to 21:00) as expected. Finally, this plot confirms that estimations from HTS data suffers from incomprehensible sparsity and irregularity behaviors since are based  on small sample of the population. Thus, even if it is a representative one, it remains insufficient to give a complete estimation on mobility flows. By contrast, using census data, leads to more regular and smooth estimations since it uses extremely large sample so that representativeness is no more needed.

\subsection{Relevant features}

In order to detect the relevant features in the learning process, we train our data with standard state-of-the-art linear model. In fact, in this way we get a proxy to the importance of each variables using its corresponding weights in a linear model. These weights provide a wealth of information on the role played by the variables. We have noticed that the SVM with linear kernel performs the closest classification rate to our best results. This enable us to identify the most relevant features linked to how people travel, using the weights of each feature relatively to the predicted variables. The learning process has an effect on the weight assigned to each variable. When using linear models, this weights provide a wealth of information on the role played by the variables. We propose in Table~\ref{table:relevant_features:workers},~\ref{table:relevant_features:students} an overview of the distribution of these weights into the four predictor categories ($HA$ represents the home attributes, $WA$ and $SA$ the workplace and school location attributes, $ATT$ the average travel time ones and $PA$ the personal attributes). We compute the sum of the features weights of a category and divided by total weights sum. This gives a quick overview of the relevance of each category for a specific variable. It can be point out in Table~\ref{table:relevant_features:workers},~\ref{table:relevant_features:students} that the four categories are relatively predictive in our model.

\begin{table}[h]
\caption{Worker most relevant variables with respect to feature weight}
\label{table:relevant_features:workers}
\begin{center}
    \begin{tabular}{|c|c|c|c|c|c|}
    \hline
    \textbf{Targets}                 &\textbf{\#1}      &\textbf{\#2}      &\textbf{\#3}       &\textbf{\#4}       &\textbf{\#5}       \\
    \hline
    \textbf{ATaW}        & 1.3 (1) & 1.0 (2) & 0.8 (3)  & 0.7 (4)  & 0.7 (5)  \\
    \hline
    \textbf{ATaH}        & 1.1 (3) & 1.1 (5) & 1.1 (4)  & 1.1 (6)  & 1.0 (7)  \\
    \hline
    \textbf{TTtW}        & 4.3 (8) & 3.4 (9) & 0.8 (10) & 0.7 (11) & 0.7 (1)  \\
    \hline
    \textbf{TTtH}        & 3.0 (8) & 2.4 (9) & 0.7 (10) & 0.6 (12) & 0.5 (13) \\
    \hline
    \end{tabular}
\end{center}
\begin{center}
    \begin{tabular}{|c|l|}
    \hline
    \textbf{Variable}        & \textbf{Description} \\
    \hline
    (1) (PA)        & Worker age \\
    \hline
    (2) (PA)        & Employment status  \\
    \hline
    (3) (PA)        & Executive or intellectual profession \\
    \hline
    (4) (PA)        & Part time work \\
    \hline
    (5) (PA)        & Full time work \\
    \hline
    (6) (PA)        & Intermediate profession \\
    \hline
    (7) (PA)        & Public transport user  \\
    \hline
    (8) (ATT)       & Average travel time on road (home to workplace)\\
    \hline
    (9) (ATT)       & Average travel time on road (workplace to home)\\
    \hline
    (10) (HA)       & Number of household with the reference person is farmer\\
    \hline
    (11) (HA)       & Number of person between 15-24 year old who are farmer\\
    \hline
    (12) (WA)       & \multirow{2}{22em}{Number of house build as primary residence between 1990 and 2004}\\
    (13) (HA)       & \\
    \hline
    \end{tabular}
\end{center}
\end{table}

The Table~\ref{table:relevant_features:workers} presents, for each predicted variable, the five most important predictor variables. The upper part of the table shows that the most determining factor for $ATaW$ is first the age of the people, but also the kind of job and if it is a part-time or a full-time work. When considering the arrival time at home, we can see that the use of a public transport is also an essential part. We will now analyze what are the key issues that determine the travel times. The table~\ref{table:relevant_features:workers} shows that the theoretical travel time (variables 7 and 8) is the most determining. Of course it is not surprising, but we can notice that actual travel times are also determined by other parameters, including the presence of farmers in the household (variables 10 and 11). In other words, these results show how rurality influences the travel time.

In Table~\ref{table:relevant_features:students}, we note that the predictor variables travel time using personal vehicle (1), age (2) and day of the week (9) are among the most determining. This is expected since the first variable explains the effects of traffic congestion on mobility patterns. Moreover, the relevance of the feature age suggests that this is a way for the model to approximate the student grade: a missing information in the predictor table. The day of week (9) is also an expected relevant feature in France since the study time schedule depends on the day of week and especially in Wednesday since the academic course hours rarely exceed half a day. It is interesting to note that the other determining features can be grouped into two main sets. First, variables (5) and (7) show how populations size and urbanization are determining in the learning process. Second, variables (3,4,6,8) and (10) are some general indicators describing the economic situation of a region: the number of senior executives, the percentage of people who actively participate in the economy, the percentage who are currently employed, and the percentage who are unemployed. They are also indicating the presence of mothers who are homemakers.

\begin{table}[h]
\caption{Student most relevant variables with respect to feature weight}
\label{table:relevant_features:students}
\begin{center}
    \begin{tabular}{|c|c|c|c|c|}
    \hline
    \textbf{Targets} & \textbf{\#1} & \textbf{\#2} & \textbf{\#3} & \textbf{\#4}\\
    \hline
    \textbf{ATaS} & 0.96 (3) & 0.68 (5) & 0.66 (6) & 0.63 (7) \\
    \hline
    \textbf{ATaH} & 0.87 (2) & 0.67 (9) & 0.48 (10) & 0.47 (8) \\
    \hline
    \textbf{TTtS} & 2.73 (1) & 2.47 (2) & 0.85 (3) & 0.70 (4) \\
    \hline
    \textbf{TTtH} & 1.58 (1) & 1.10 (2) & 0.86 (3) & 0.81 (8) \\
    \hline
    \end{tabular}
\end{center}
\begin{center}
    \begin{tabular}{|c|l|} 
    \hline
    \textbf{Variable}        & \textbf{Description} \\
    \hline
    (1) (ATT) & Travel time using personal vehicle\\
    \hline
    (2) (PA) & Student's age\\
    \hline
    (3) (HA) & Number of women workers older than 15 years\\
    \hline
    (4) (SA) & Number of active workers between 15-64 years old\\
    \hline
    (5) (HA) & Number of people older than 55 years\\
    \hline
    (6) (SA) & Number of senior executives older than 55 years\\
    \hline
    (7) (HA) & Number of dwellings\\
    \hline
    (8) (HA) & Number of companies\\
    \hline
    (9) (PA) & Day of the week: Wednesday\\
    \hline
    (10) (SA) & Number of unemployed people older than 55 years\\
    \hline
    \end{tabular}
\end{center}
\end{table}

In other words, all these observations concert with the previous results in \cite{gonzalez2008understanding,arai2014understanding,lenormand2015influence,louf2014scaling,louf2014congestion}. It is suggesting that human mobility patterns result from an interaction of socio-demographic attributes of individuals with socio-economic indicators of cities and that this phenomenon is similar in all the countries.

\subsection{Correlation with other mobility data}

Our model present some good results, and we develop an interest in comparing the data obtained with other real mobility data. For that we propose to use O-D data estimations from one of the main mobile network operator of France. It had been shown that mobile network data are a good indicator of Human mobility \cite{smoreda2013spatiotemporal}. Their Floating Mobile Data (FMD) corresponds to a matrix containing O-D estimations observed over the same studied area, over a similar period. The mobile network operator keeps trips whose immobility duration on destination greater than 30 minutes. Their dataset should include our daily work/study round-trips. Moreover, the estimations have been adjusted to population densities, i.e. this dataset contains close-to-reality data. We propose to study the statistical link between our estimations and this dataset.

\begin{figure}[h]
   \centering
   \includegraphics{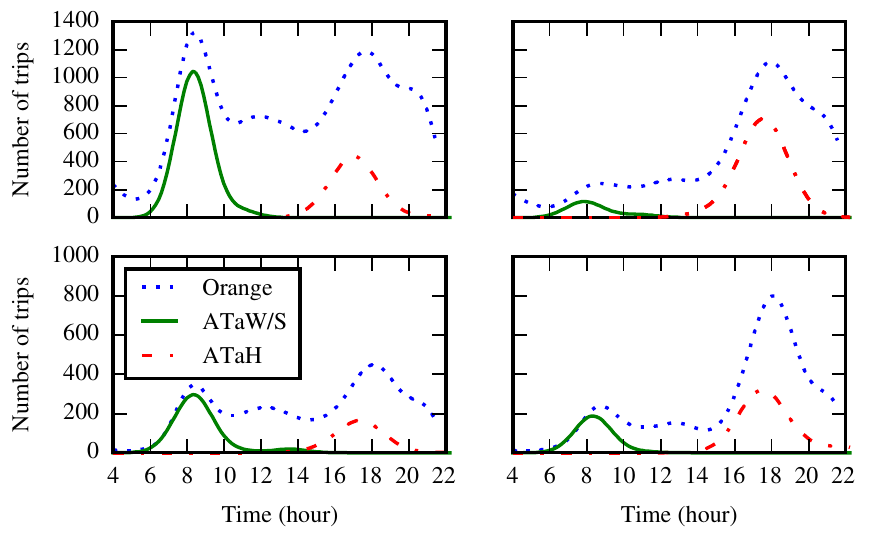}
   \caption{Example of 4 origin-destination trips distribution}
   \label{figure:orange_atah_ataw}
\end{figure}

We present in Fig.~\ref{figure:orange_atah_ataw} the comparison between the displacements observed by the network operator and the displacements estimated by our model over a same O-D trip. The four graphs displayed correspond to four different O-D trips between districts of Paris. The displacements estimated by our model are furthermore characterized with a trip reason which can be work/school (ATaW/S) or home (ATaH).\\

\begin{figure}[h]
   \centering
   \includegraphics{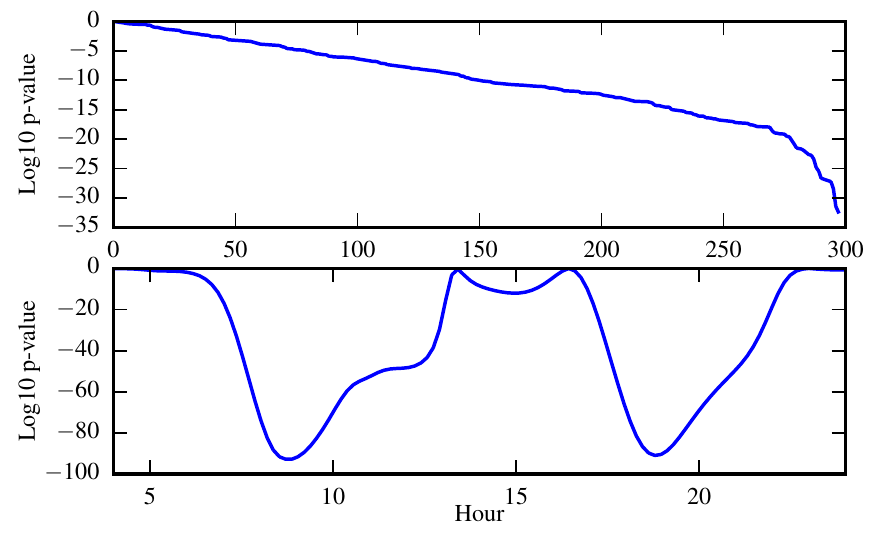}
   \caption{The sorted p-values over 296 O-D in Paris (top) and the p-values for a segmented day with 2 vectors which contain the 296 O-D (bottom) (false discovery rate corrected p-values)}
   \label{figure:p_value}
\end{figure}

In Fig.~\ref{figure:p_value} we compute the p-value for testing non-correlation between the network operator dataset and our estimated data, for 296 origin-destination trips. The results show a statistically significant correlation between the two data sources (almost 92.9\% in the majority).
This confirm that home, work and study related trips constitute an important and constant part of Human mobility. In order to refine these results we also study the amplitude of this correlation between the two O-D matrices for each time slot over the day. The results displayed in Fig.~\ref{figure:p_value} allow us to conclude that the work related flows estimations are strongly correlated to classical commuting periods (08:00 to 10:00 and 16:00 to 21:00).

\section{Conclusion}
\label{section:conclusion}

The purpose of this experimentation was to develop a multi-task learning model. The state-of-the-art led us to use a multi-task neural network based model. The proposed approach was applied in a real context on eleven urban communities. We compared our approach with simple-task Multi-Layer Perceptron neural network and we showed a more stable and accurate classification with a multi-task approach. This approach leads to an estimation of origin-destination matrices with census data. This allows to fuse it with other mobility sources to get a complete and stable estimation of human mobility flows. The estimated origin-destination matrices are in a good agreement with origin-destination matrices coming from actual independent Household Travel Survey data and Floating Mobile Data.

Future works focus on the improvement of the classification rate by using a deeper network. Moreover, we do believe that unsupervised pre-training (as Restricted Boltzmann machines and auto-encoders), may be used to avoid overfitting of such deep networks. This will also allow to use unlabelled data. The heterogeneity of mobility data make this task more challenging. In fact, generally, such pre-training techniques suppose that the predictor features are all homogeneous (binary, Gaussian...). In our case, we need deal with binary, continuous, and nominal variables at the same time which is more difficult.

\begin{acknowledgements}
The authors would like to thank both CEREMA (provider), ADISP-CMH (distributor) for the $HTS$ data, the ANR for granting the project Norm-Atis under Grant ANR-13-TDMO-07 and Orange Fluxvision \cite{orange2018fluxvision}.
\end{acknowledgements}

\bibliographystyle{spmpsci}
\bibliography{main}

\end{document}